\pdfoutput=1

\documentclass[11pt]{article}

\usepackage{ACL2023}
\usepackage{svg}

\usepackage{subcaption}
\usepackage{caption}
\usepackage{times}
\usepackage{latexsym}
\usepackage{hyperref}
\usepackage{multirow} 
\usepackage{tabularx}
\usepackage{booktabs}
\usepackage{graphicx}
\usepackage{booktabs} %

\usepackage[T1]{fontenc}

\usepackage[utf8]{inputenc}
\usepackage{graphics}
\usepackage{microtype}

\usepackage{inconsolata}
\usepackage{amsmath} 

\usepackage{soul}

\title{Towards Detecting Persuasion on Social Media: \\From Model Development to Insights on Persuasion Strategies}

\author{
    Elyas Meguellati$^1$,
    Stefano Civelli$^1$,
    Pietro Bernardelle$^1$,
    Shazia Sadiq$^1$,
    Irwin King$^2$,
    and Gianluca Demartini$^1$\\[0.5em] 
    $^1$University of Queensland, Brisbane, Australia\\
    $^2$The Chinese University of Hong Kong, Hong Kong SAR, China\\[1em] 
    \texttt{$^1$\{m.meguellati, s.civelli, p.bernardelle, s.sadiq, g.demartini\}@uq.edu.au}\\
    \texttt{$^2$\{king@cse.cuhk.edu.hk\}} 
}

\begin{document}
\maketitle

\begin{abstract}
Political advertising plays a pivotal role in shaping public opinion and influencing electoral outcomes, often through subtle persuasive techniques embedded in broader propaganda strategies. Detecting these persuasive elements is crucial for enhancing voter awareness and ensuring transparency in democratic processes. This paper presents an integrated approach that bridges model development and real-world application through two interconnected studies. First, we introduce a lightweight model for persuasive text detection that achieves state-of-the-art performance in Subtask 3 of SemEval 2023 Task 3 while requiring significantly fewer computational resources and training data than existing methods. Second, we demonstrate the model’s practical utility by collecting the Australian Federal Election 2022 Facebook Ads (APA22) dataset, partially annotating a subset for persuasion, and fine-tuning the model to adapt from mainstream news to social media content. We then apply the fine-tuned model to label the remainder of the APA22 dataset, revealing distinct patterns in how political campaigns leverage persuasion through different funding strategies, word choices, demographic targeting, and temporal shifts in persuasion intensity as election day approaches. Our findings not only underscore the necessity of domain-specific modeling for analyzing persuasion on social media but also show how uncovering these strategies can enhance transparency, inform voters, and promote accountability in digital campaigns.
\end{abstract}

\section{Introduction}
The spread of information through digital platforms brings to light the practice of propaganda, which aims to shape public opinion to foster certain agendas. Propaganda can be observed across various media, making it challenging for information consumers to recognize, especially for those new to its nuances. Its effectiveness lies in its ability to blend seamlessly into the content, often requiring a keen eye to detect its subtle cues. Furthermore, the increasing amount of information on social media adds to the challenges for experts to manually sort through the content in order to spot and flag elements of propaganda \citep{glowacki2018news,Tardaguila2018FakeNews}.
The detection of propaganda has become a critical area of study, particularly with the rising concern over misleading information being spread at scale online.

\begin{figure}[tbp]
\includegraphics[width=\columnwidth]{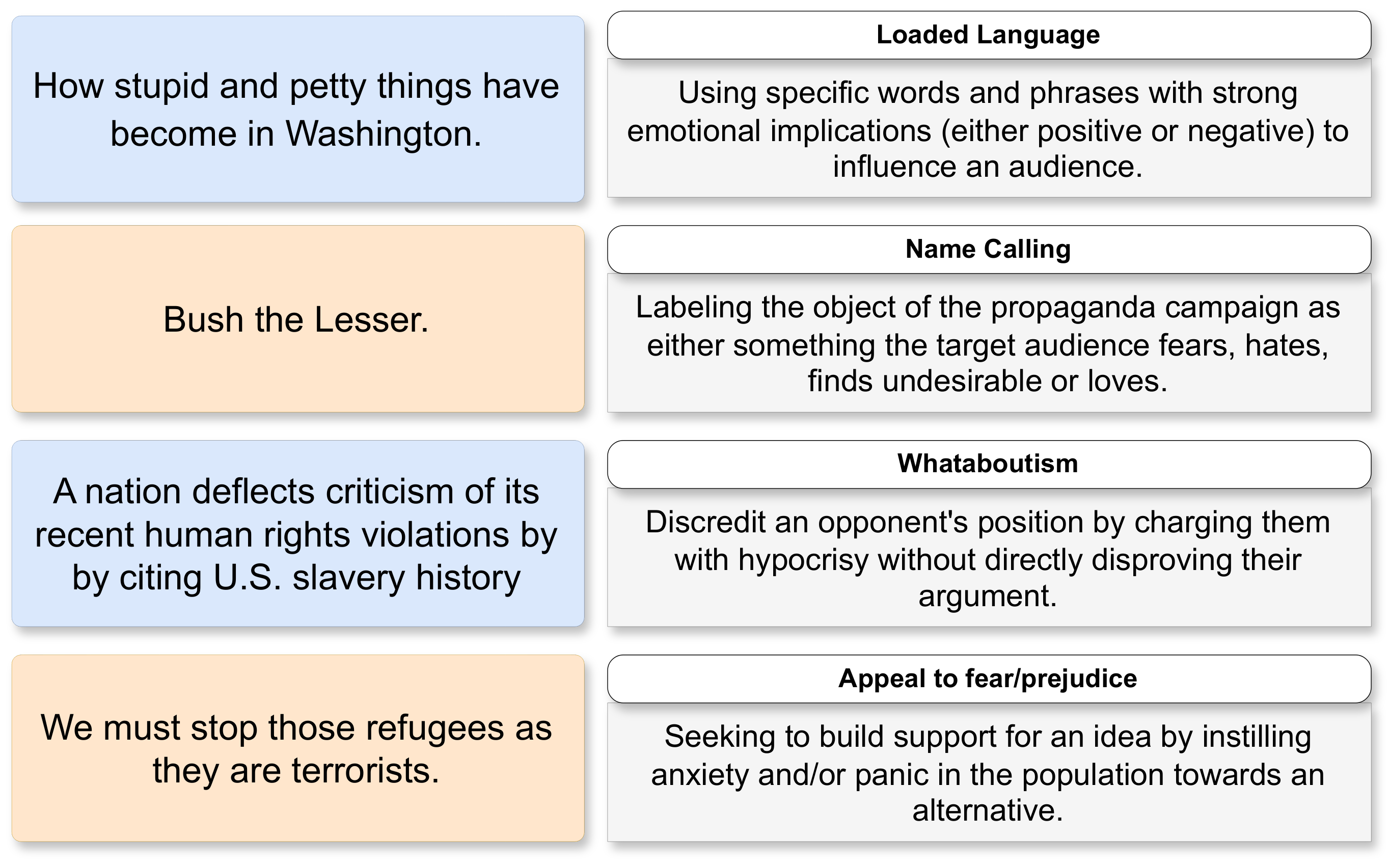}
  \caption{Examples of Persuasion techniques from the SemEval dataset with explanation of the technique.}
  \label{fig:persuasion_example}
\end{figure}

In this paper we focus on persuasive content, a key element of propaganda. Persuasion is about influencing people's beliefs, attitudes, or actions through dialogue and reasoning. It is a process where communicators seek to convince others on certain issues, respecting their freedom to choose \citep{Perloff2017Dynamics}. Incidents like Cambridge Analytica  have shown how persuasive tactics can be amplified on social media, using data to tailor messages for political ends \citep{boerboom2020cambridge}. Figure \ref{fig:persuasion_example}  illustrates various persuasive techniques employed in language to elicit distinct emotional responses\footnote{\url{https://propaganda.math.unipd.it/semeval2024task4/definitions22.html}}.

Addressing the need for nuanced persuasion detection, SemEval-2023 Task 3 organizers introduced Subtask 3, focusing on categorizing 23 persuasive techniques in articles. Our contributions include: (1) Developing a novel, more efficient, multi-label classification model for persuasion detection with reduced data requirements and a lighter transformer. (2) Curating and partially annotating a large dataset of political ads published during the 2022 Australian Federal Election on Facebook. (3) On this dataset, our model provides new insights into persuasion strategies in political advertising, demonstrating its application in practice.

\section{Related Work}
In the domain of text analysis, the distinction between the detection of persuasion and propaganda has gradually become more defined. Historically, these domains were often conflated, given the emphasis on the identification and categorization of rhetorical techniques such as loaded language, name-calling, and emotional appeals. Early contributions, such as the work by \citet{rashkin2017}, centered on document-level classification, categorizing texts into broad groups including trusted, satire, hoax, and propaganda. \citet{barron2019} extended this by developing a corpus to differentiate propagandistic from non-propagandistic content, analyzing the impact of writing style and readability on predicting the source of the article.

The granularity of the classification increased with subsequent studies. \citet{habernal2017}, who focus on persuasion,  annotated arguments with five distinct logical fallacies. This area of inquiry was further expanded by \citet{dasmartino2019}, who introduced a corpus annotated with an extensive set of propaganda techniques and developed advanced models for  span detection and classification. Their efforts were extended in later work that addressed the limitations of transformers \citep{chernyavskiy2021transformers} and explored interpretable propaganda detection \citep{yu2021interpretable}. Additional studies have examined propaganda techniques in diverse contexts, such as memes \citep{dimitrov2021detecting} and code-switched text \citep{salman2023detecting}, as well as the interplay between propaganda and social coordination \citep{hristakieva2022spread} and the spread of COVID-19 related propaganda on social media \citep{nakov2021covid}.

The introduction of SemEval-2023 Task 3 Subtask 3, as presented by \citet{piskorski2023multilingual}, marked a significant shift by explicitly assigning persuasion detection as a distinct subtask within the broader context of propaganda classification, alongside two other subtasks: (1) news genre categorization (objective reporting vs. opinion vs. satire) and (2) framing detection (highlighting specific aspects of content). The persuasion detection subtask leveraged a comprehensive multilingual dataset comprising of 1,612 news articles, carefully annotated with 23 persuasion techniques (i.e., a multi-label classification task), thus expanding the scope of media content computational analysis. The used baseline was a \textit{linear SVM} trained using word \textit{uni-grams} and \textit{bi-grams} \cite{piskorski2023semeval}.

Within this refined framework, innovative approaches recently emerged, heavily relying on transformers and data augmentation, without significant emphasis on feature engineering. \citet{purificato2023} combined an ensemble of fine-tuned transformer models, applying a weighted average to predictions for English to improve sensitivity to different persuasion techniques. Meanwhile, \citet{Hromadka2023KinitveraAI} attained state-of-the-art (SOTA) results through minimal preprocessing and fine-tuning of the XLM-RoBERTa-large model \cite{DBLP:journals/corr/abs-1911-02116}. A key aspect of their model's performance was the focus on calibrating the confidence threshold.


As compared to the existing literature, we introduce a lightweight transformer model that exhibits high levels of effectiveness on the persuasion detection subtask of SemEval-2023. Our model achieves comparable results to other top-performing models, such as APatt \citep{purificato2023} and KInIT \citep{Hromadka2023KinitveraAI}, while focusing on data and computational efficiency. Notably, our approach requires approximately 86.8\% less data, up to 77.6\% fewer parameters than APatt, and 86.2\% less training time as compared to KInIT. This work extends the body of knowledge on persuasion detection in digital media, also providing new insights into persuasive strategies within online political advertising.

\section{Study 1: Persuasion as a Multi-Label Classification for SemEval-23}

\subsection{Dataset Description}
The SemEval-23 Task 3 dataset is a comprehensive multilingual corpus collected from various globally discussed topics, including the COVID-19 pandemic, abortion-related legislation, migration, and the Russo-Ukrainian conflict, among others. The dataset includes articles in nine languages: English, French, German, Georgian, Greek, Italian, Polish, Russian, and Spanish published between 2020 and mid-2022. The articles were sourced from both mainstream and alternative media to capture a wide range of perspectives. Each article was annotated for persuasive techniques at the span level, utilizing a detailed taxonomy designed for this task. Approximately 40 annotators, proficient in their respective languages, ensured a high-quality annotation of the dataset. See Table~\ref{tab:dataet_semeval_stats} in the Appendix.

\subsection{Methodology}
The SemEval-23 Task 3 Subtask 3 presents a multi-label classification task with imbalanced labels across 23 persuasion techniques, withholding the test labels and requiring participants to submit the predictions to an online platform for evaluation. 
For this task, we introduce our novel PPAsy (Persuasion Prediction Asymmetric) method, which utilizes an asymmetric binary cross-entropy loss function to effectively manage label imbalance. 
The official metric set by the organizers to assess performance is the micro-averaged F1 score (\textit{F1-micro}), complemented by the F1-macro score.

\paragraph{Data Preparation.}
We translated all non-English training documents into English using the Google API\footnote{\url{https://cloud.google.com/translate}} machine translation model to standardize the data for uniform model training. 
We trained our model on individual datasets, each translated from the source language to English. Additionally, we trained the model on a combined dataset, which included the translated training data from all languages along with the original English dataset. The French language dataset, comprising of 211 documents out of a total of 1,612, exhibits a notably high average number of persuasion techniques per document. It contains all 23 labels, in contrast to the English dataset, which includes only 19. Consequently, the French-to-English was the only dataset employed to train our model which was subsequently utilized to generate predictions for the English test set.

\paragraph{Preprocessing.}

We employed minimal preprocessing for all translated datasets, consisting only of lemmatization and lowercasing using SpaCy\footnote{\url{https://spacy.io/}}. We specifically considered the role of punctuation, informed by its demonstrated stylistic significance in textual analysis \citep{{darmon2021pull}}. Noting that sentences including punctuation yielded better results, as shown in Table \ref{tab:ablation_study}, we opted for keeping it. This outcome suggests a potential influence of punctuation on the model's ability to interpret text.

\paragraph{Model Selection and Training.}

In our exploration of various machine learning models, including SVM, CNN, and XLNet, we ultimately selected XLNet-base \cite{DBLP:journals/corr/abs-1906-08237} for its  balance between performance and resource efficiency. This choice was influenced by the model's relatively lower computational demands as compared to the ensembles of large transformer models and extensive data augmentation with multiple languages used by other SOTA models for persuasion detection such as APatt and KInIT~\citep{purificato2023,Hromadka2023KinitveraAI}.

\paragraph{Loss Function.}

To address the challenges inherent in classification with many classes and skewed label distribution, we opted for a custom adaptation of the binary cross-entropy loss function. Our implementation involves asymmetric weighting to accommodate for class imbalance:

\begin{gather*}
L(y, \hat{y}) = \frac{1}{N} \sum_{i=1}^{N} w_i \cdot BCE(y_i, \hat{y}_i), \\
\text{where } w_i = \beta \cdot  y_i+ (1 - \beta) \cdot (1 - y_i)
\end{gather*}

In this formulation, \(y\) denotes the true labels, and \(\hat{y}\) represents the predicted probabilities. \(BCE\) refers to the binary cross-entropy, and \(w_i\) is the instance-specific weight. The \(\beta\) parameter is employed to modulate the loss contributions from the positive and negative classes, with the aim of mitigating label imbalance effects by variably penalizing misclassifications.

\begin{table*}[ht]
\centering
\small
\begin{tabular}{lcccc}
\toprule
\textbf{Iteration} & \textbf{Component Removed} & \textbf{F1 Micro (\%)} & \textbf{F1 Macro (\%)} & \textbf{Comparison to SOTA} \\
\midrule
SOTA & Full model & 40.62 & 21.78 & N/A \\
1 & Translation & 28.35 & 3.4 & -12.27 \\
2 & Punctuation & 39.71 & 19.65 & -0.91 \\
3 & Loss function & 33.94 & 12.69 & -6.68 \\
4 & Epoch (non-optimal) & 38.49 & 16.42 & -2.13 \\
\bottomrule
\end{tabular}
\caption{Ablation study results show the impact of component removal. Translation, utilizing the Google API—an approach widely adopted by SemEval participants—results in the largest performance drop. Conversely, our novel loss function substantially boosts F1 scores, highlighting its unique contribution.}
\label{tab:ablation_study}
\end{table*}

\paragraph{Prediction Threshold Calibration.}

\begin{figure}[tbp]
  \hspace*{-0.5cm} 
  \includegraphics[width=\columnwidth]{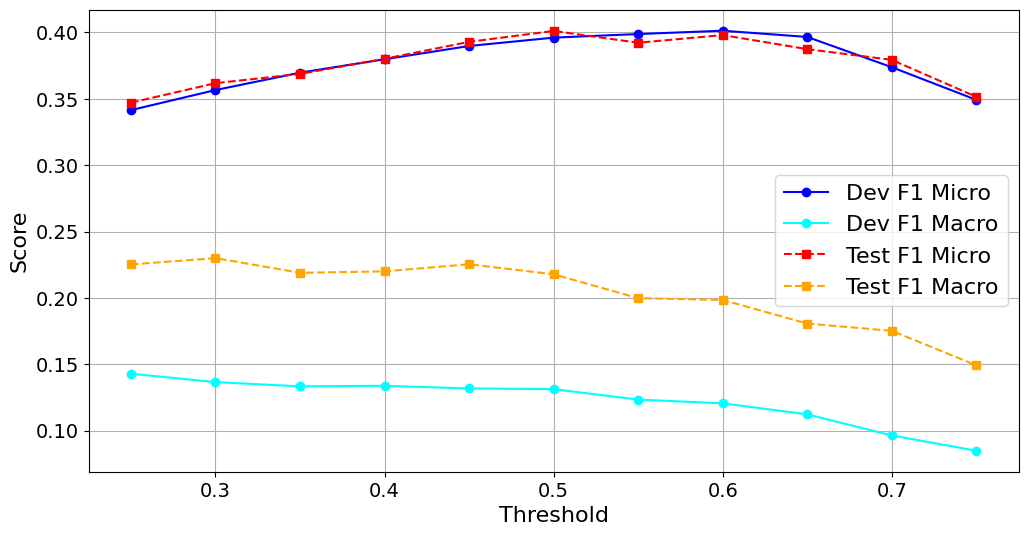}
  \caption{Model Performance across Different Thresholds.}
  \label{fig:model_performance}
\end{figure}

When converting model output probabilities into binary class predictions, selecting an optimal threshold is crucial. However, as shown in Table~\ref{tab:dataet_semeval_stats}, there are significant discrepancies in average persuasion (AVG pt) between the English development (Dev) and test sets. This difference suggests that improvements observed in the Dev set may not consistently translate to comparable improvements in the Test set, and vice versa. To strike a balance between optimizing F1-micro and F1-macro performance, we maintained a default threshold of 0.5, as illustrated in Figure \ref{fig:model_performance}, despite the Dev set having an optimal threshold of 0.6.

\begin{table*}[ht]
\centering
\small
\begin{tabular}{@{}lccccc@{}}
\toprule
\textbf{Method} & \textbf{Dataset Size} & \textbf{Configuration} & \textbf{Parameters} & \textbf{F1-Micro (\%)} & \textbf{F1-Macro (\%)} \\
\midrule
PPAsy-Conv (Ours) & 153K & CNN & N/A & 35.01 & 8.89 \\
APatt \citep{purificato2023} & 1.16M & Transformers Ensemble & 492M & 37.56 & 12.92 \\
KInIT \citep{Hromadka2023KinitveraAI} & 1.16M & RoBERTa-large & 355M & 40.38 & 15.86 \\
\textbf{PPAsy-XLNet} (Ours) & 153K & XLNet-base & 110M & 40.62 & 21.78 \\
\midrule
Baseline \cite{piskorski2023semeval} & 469K & SVM & N/A & 19.51 & 6.93 \\
\bottomrule
\end{tabular}
\caption{Performance metrics of models on the English test set, illustrating PPAsy's efficiency with reduced training data and lower computational complexity. Notably, PPAsy-XLNet outperforms other transformers with fewer parameters and training volume. The SVM configuration is the baseline set by SemEval-23 \cite{piskorski2023semeval}.}
\label{tab:model_performance_subtask3}
\end{table*}

\paragraph{Novel Contribution.}
Our approach builds upon established approaches such as \textit{focal loss} \cite{lin2017focal} and  \textit{conceptually class-balanced loss} \cite{cui2019class}, which asymmetrically adjust the loss function based on class sample sizes. In addition, we have adapted the binary cross-entropy framework by integrating an external class-specific weighting mechanism, facilitated by a parameter \(\beta\) to modulate weights \(w_i\) for each instance. This adaptation retains the  simplicity of binary cross-entropy while addressing class imbalance. While this approach is consistent with the insights presented by \citet{teimas2023detecting}, it applies an unconventional loss function which substantially improves the performance in this multi-label classification task with skewed label distributions as shown by Iteration 3 in Table \ref{tab:ablation_study}.
Our implementation is available online as open-source at [redacted-for-blind-review].

\subsection{Findings and Implications}

\begin{table*}[t]
\centering
\small\resizebox{\textwidth}{!}{%
\begin{tabular}{>{\centering\arraybackslash}m{3cm}|r|r|r|r|r|r|r|r}
\hline
\textbf{Model} & \multicolumn{2}{c|}{\textbf{Training Time (epoch)}} & \multicolumn{2}{c|}{\textbf{Validation Time}} & \multicolumn{2}{c|}{\textbf{F1 Micro (\%)}} & \multicolumn{2}{c}{\textbf{F1 Macro (\%)}} \\
    & \textbf{French} & \textbf{All} & \textbf{French} & \textbf{All} & \textbf{French} & \textbf{All} & \textbf{French} & \textbf{All} \\
\hline
XLNet-Base & 136 (27) & 492 (153) & 10 & 34 & 40.62 & 39.71 & 21.78 & 23.16 \\
XLNet-Large & 305 (76) & 623 (311) & 29 & 99 & 36.78 & 36.75 & 7.92 & 6.64 \\
RoBERTa-Base & 128 (25) & 428 (128) & 11 & 32 & 40.38 & 41.12 & 11.69 & 21.59 \\
RoBERTa-Large & 277 (69) & 988 (277) & 27 & 90 & 34.18 & 34.17 & 4.01 & 4.13 \\
DistilBERT & 100 (20) & 399 (100) & 8 & 27 & 40.60 & 36.50 & 20.62 & 19.14 \\
\midrule 
XLM-RoBERTa-Base & 111 (22) & 364 (111) & 9 & 30 & 31.08 & 32.26 & 5.73 & 6.21 \\
XLM-RoBERTa-Large & 281 (56) & 432 (281) & 24 & 80 & 34.18 & 24.68 & 4.01 & 2.06 \\
\hline
\end{tabular}%
}
\caption{Overview of model training and validation performance across various configurations. Training and validation times are reported in seconds, with the values in parentheses indicating the time per epoch. The "French" and "All" columns represent models trained on French or all languages translated to English, respectively, except for XLM-RoBERTa, where no translation was performed. All models were validated and tested on the English dataset.}
\label{tab:models_performance}
\end{table*}

\paragraph{Data Efficiency.} We benchmarked the performance of our PPAsy models against  published models which are highly effective on the SemEval-23 task. Our PPAsy-XLNet, utilizing the base version of XLNet with 110M parameters, achieved an F1-Micro score of 40.62\% and an F1-Macro score of 21.78\%, as shown in Table \ref{tab:model_performance_subtask3}. These results are particularly significant considering the model's operation on a dataset size of only 153K tokens, substantially smaller than the 1.16M tokens used by both APatt and KInIT ~\citep{purificato2023,Hromadka2023KinitveraAI}.

\paragraph{Computational Efficiency.} A comparative analysis reveals further efficiencies of our PPAsy-XLNet model. APatt employs an ensemble approach, integrating five transformer models: XLNet, RoBERTa, BERT, ALBERT, and DeBERTa, with respective parameters of 110M, 125M, 110M, 12M, and 135M, for a total of 492M parameters. Despite the extensive parameter count, APatt's ensemble achieves F1-Micro and F1-Macro scores of 37.56\% and 12.92\%, respectively. In contrast, KInIT's methodology leverages a fine-tuned RoBERTa-large model with 355M parameters, employing layer freezing techniques to explore various fine-tuning strategies based on the portion of the model that is frozen, ultimately achieving F1-Micro and F1-Macro scores of 40.38\% and 15.86\%, respectively. Both previous approaches utilize all the labeled datasets from other languages to augment their training sets to boost their models' performance while our approach only uses the French subset translated to English.

Despite the considerable computational resources and the augmented training data used by most Semeval-23 participants, our PPAsy-XLNet model, operating with significantly fewer parameters and a smaller, focused training dataset, outperforms these approaches while reducing the need for data and computational resources. 
The French-to-English training dataset, instantiating all 23 labels and exhibiting the highest average of persuasion techniques per document, has notably contributed to the robustness of our model training. This dataset supports the superior performance metrics achieved by  PPAsy-XLNet in our paper (see Table  \ref{fig:model_performance}). PPAsy-XLNet achieves an F1-Micro of 40.62\% and an F1-Macro of 21.78\%, outperforming the baseline models. As shown in Table \ref{tab:models_performance}, DistilBERT obtained the shortest execution times, the efficiency margins between it and PPAsy-XLNet are minimal. Notably,  KInIT \citep{Hromadka2023KinitveraAI}, leveraging a RoBERTa-Large configuration, demands considerably more computation time (988 seconds) as compared to our PPAsy-XLNet (136 seconds). These results along with the model performance comparison confirm the advantage of our model in resource-constrained scenarios. For details on the computational resources used during our experiments, see Appendix \ref{sec:comp_resources}.

\section{Study 2: Persuasion in the 2022 Australian Federal Election Campaign on Social Media Political Advertisement}
This second study extends the exploration of the PPAsy-XLNet model introduced in Study 1 to  persuasion detection in political advertisement, using as benchmark the 2022 Australian federal election campaign for which we collected and partially annotated a new dataset, called APA22.
First, to investigate the domain adaptability of PPAsy-XLNet, we employed the model trained on the SemEval-23 dataset to identify persuasion on the APA22 dataset.
Subsequently, we cross-validated the model directly on the APA22 dataset in order to determine the performance difference.
Study 2 not only tests our model's applicability to real-world data, but also offers preliminary insights into the dynamics of digital persuasion in political campaigns on social media.

\subsection{The APA22 Dataset}
\paragraph{Data Collection.}
Using the Meta Ad Library API\footnote{\url{https://www.facebook.com/ads/library/}} we collected the online  advertisements related to the Australian federal election held on May 21\textsuperscript{st} 2022. Actively running advertisement campaigns were captured with a six-hour frequency between March 1\textsuperscript{st}, 2022, and June 18\textsuperscript{th}, 2022.
Each ad includes attributes such as the creation time, target demographic distributions, ad text, funding entity, ad URL, and metrics of ad impression and spending.
The resulting dataset consists of 56,958 unique ads, each containing an average of 3.86 sentences.
The APA22 dataset offers a substantial resource for investigating the dynamics and the strategies of political advertising on social media~\footnote{The dataset collection process has undergone ethics review at the authors' organization. This dataset is available online at [redacted-for-blind-review].}.

\paragraph{Data Labeling.}
A subset of 658 samples from the APA22 dataset was randomly selected and manually annotated for persuasive content.
The annotation quality was assessed by means of inter-annotator agreement. Four independent annotators annotated an overlapping subset of the labelled dataset on which they obtained an agreement rate of 0.86 Fleiss’ Kappa.

The APA22 dataset was created and annotated prior to the release of SemEval-23, leading to differences in the two annotation schemes.
To align the two datasets for comparative analysis, we simplified the annotations to a binary scheme, categorizing each sentence as either 'neutral' or 'persuasive'. This binarization process yielded 368 'persuasive' and 290 'neutral' sentences within the annotated subset of the APA22 dataset.

The data was then split into training (75\%, 493 sentences) and testing (25\%, 165 sentences)  using \textit{stratified sampling}. A \textit{fixed seed} was used to ensure consistency and reproducibility in the test set.


\begin{table}
\centering
\begin{tabular}{l@{\hskip 0.1in}c@{\hskip 0.1in}c@{\hskip 0.1in}c@{\hskip 0.1in}c}
\hline
\multirow{2}{*}{\textbf{Model}} & \multicolumn{2}{c}{\textbf{SemEval-23}} & \multicolumn{2}{c}{\textbf{APA22}} \\
\cline{2-5}
 & \textbf{Acc} & \textbf{F1} & \textbf{Acc} & \textbf{F1} \\
\hline
PPAsy-Conv & 0.55 & 0.55 & 0.79 & 0.79 \\
SVM & 0.56 & 0.55 & 0.76 & 0.75 \\
\textbf{PPAsy-XLNet} & \textbf{0.59} & \textbf{0.55} & \textbf{0.82} & \textbf{0.82} \\
\hline
\end{tabular}
\caption{Performance comparison of models trained on SemEval-23 or on APA22 evaluated on the APA22 test set for political ad persuasion detection.}
\label{tab:perofrmance_semeval_comparison}
\end{table}

\subsection{Domain Adaptability and Re-Training}

In exploring the \textit{transferability} of models across different domains, we aimed to assess how effectively a model trained on the SemEval-23 dataset, primarily composed of news articles, could adapt to the context of political social media ads as in the APA22 dataset. 

Our experimental results show a notable lack of domain adaptability when models trained on the \textit{binarized} SemEval-23 dataset were tested on the APA22 test set. The PPAsy-XLNet model, which achieved SOTA in SemEval-23, exhibited a poor performance, scoring an accuracy of 59.39\% when tested on the APA22 dataset, as shown in Table \ref{tab:perofrmance_semeval_comparison}. This discrepancy highlights the challenges in transferring model learning from a dataset sourced from news articles to the context of political ads on social media.
Conversely, when models were trained on the APA22 dataset, they obtained significantly better results (e.g., PPAsy-XLNet accuracy of 82.42\%). 
%
These findings suggest that while models have potential for high performance, their effectiveness in real-world scenarios, such as social media ad analysis, may be limited when trained on datasets not representative of the application context.

\subsection{Insights on Political Campaigning on Social Media}

\begin{table*}[ht]
\centering
\begin{tabular}{lcc}
\hline
\textbf{Attribute} & \textbf{High Persuasion} & \textbf{Low Persuasion} \\
\hline
Ads (\%) & 26,419 (46.4\%) & 7,216 (12.6\%)\\
Avg Impressions & 25,407 & 17,441 \\
Avg Ad Spending (\$) & 393.6 & 265.6 \\
Avg Ad duration & 11 days & 8 days \\
\hline
Top Funding Entity & Solutions for Australia & Liberal Party of Australia \\
\hline
\multirow{2}{*}{Top bi-grams} & "better future" & "business small" \\
& "change climate" & "health mental"\\
\hline
Top Words & government, future, change & community, local, support \\
\hline
\end{tabular}
\caption{Statistics of political ads in the APA22 dataset using our PPAsy-XLNet model predictions. High persuasion ads consist of 80\% or more persuasive sentences, while low persuasion ads contain 20\% or fewer. Top funding entities are those that spent the most on ads. Bi-grams show the most common word combinations in high and low persuasive ads. Top words are identified using tf-idf as explained in Appendix \ref{sec:tf-idf}.}
\label{tab:persuasion_ads_comparison}
\end{table*}



The following analysis aims to unravel subtleties of the language adopted in high and low persuasion ads. PPAsy-XLNet, trained on the manually annotated portion of the APA22 dataset, was used to label each sentence in each APA22 advertisement as either 'neutral' or 'persuasive'.


Based on the percentage of persuasive sentences contained in each ad, we define the concepts of high and low persuasive ads. 
In our dataset, 26,419 advertisements (46.4\% of the total) were found to be highly persuasive (i.e., more than 80\% of the sentences in the advertisement considered persuasive). On the other hand, 7,216 ads (12.6\% of the total) were categorized as having low levels of persuasion, meaning that 20\% or fewer sentences were classified as persuasive.


\paragraph{Persuasive Content Dynamics.}
Table \ref{tab:persuasion_ads_comparison} offers insights into the persuasion found in the APA22 dataset. We observed that low persuasion ads tend to obtain fewer impressions, averaging 17,441, and incur lower spending, with an average of \$265.6. In contrast, ads categorized as highly persuasive achieve higher averages, with 25,407 impressions (45.6\% more) and \$393.6 in spending (48.2\% more), though both target the \textit{same} age and gender demographics. These findings suggest a potential correlation between the persuasive intensity of ads and their campaign reach and spend, with highly persuasive ad campaigns lasting on average 11 days compared to 8 days for low persuasion ones. To gain a better understanding of what exactly makes an ad persuasive, we next compare patterns in the language used in high and low persuasion ads.

\paragraph{Bi-grams Analysis.}
We analyze bi-gram frequencies to discern notable differences in messaging strategies.
Bi-grams commonly found in highly persuasive ads, such as 'climate change' and 'better future' (see Table \ref{tab:persuasion_ads_comparison}) seem to reflect an emphasis on policy and future-oriented values.
This might suggest an attempt to appeal to the audience's aspirations and ongoing concerns.
An illustrative example of this persuasive approach is the sentence, \textit{``Don’t risk more Morrison. Only a vote for Labor will kick out this government and deliver a better future."} This ad employs emotive language implying negative consequences (``Don't risk") and incorporates a strong call to action (``vote for Labor").
References to political figures and government entities, like frequent bi-grams "government labor", "scott morrison", and related variations, indicate targeted political messaging. 
It is noteworthy that although this type of bi-grams is also present in text classified as low in persuasion, their prevalence is significantly reduced.




Conversely, low persuasion ads feature more informational and less emotionally charged bi-grams. These bi-grams are often descriptive, focusing on geographical, cultural, or administrative entities and typically lack a compelling call to action or emotional appeal. The language in these ads tends to focus on identity, locality, or administrative details, presenting information in a straightforward manner without the dynamic or actionable language typical of persuasive ads. For instance: \textit{``The small business tax rate has been reduced to 25\% – the lowest level in 50 years. 4,000 local small businesses in Cowan will be able to access our new 20\% bonus deduction for upskilling their staff."} This statement illustrates how low persuasion ads usually provide factual information relevant to a specific audience without aggressive persuasion or emotional appeals. For visual examples of these advertising strategies, see Figure \ref{fig:high-persuasive-ad} and Figure \ref{fig:low-persuasive-ad} in the Appendix.

\paragraph{TF-IDF Analysis.}



A TF-IDF analysis reveals potential differences between the lexical choices in highly persuasive ads versus those that are low persuasion. 
Highly persuasive ads appear to use words that evoke broader national themes and encourage immediate action, such as "vote", "government", "future", "need", "better", and "plan". These words suggest that the ads may employ a strong, action-oriented vocabulary, potentially tapping into nationwide concerns, which could contribute to their persuasiveness.
Conversely, words prevalent in less persuasive ads, such as "local", "support", and "team" suggest a more localized or specific focus. These terms might not provide a sense of urgency, potentially making these advertisements less engaging.
Words like "community", "vote", "Australia", and "government" appear equally in both types of ads, indicating their general importance in Australian political advertising.
Further research is needed to establish a definitive link between specific words and their persuasion effect. Nonetheless, these initial observations provide a foundation for future studies examining the relationship between language and persuasion in political advertising.


 
\paragraph{Temporal Dynamics.}

\begin{figure}[tbp]
  \includegraphics[width=\columnwidth]{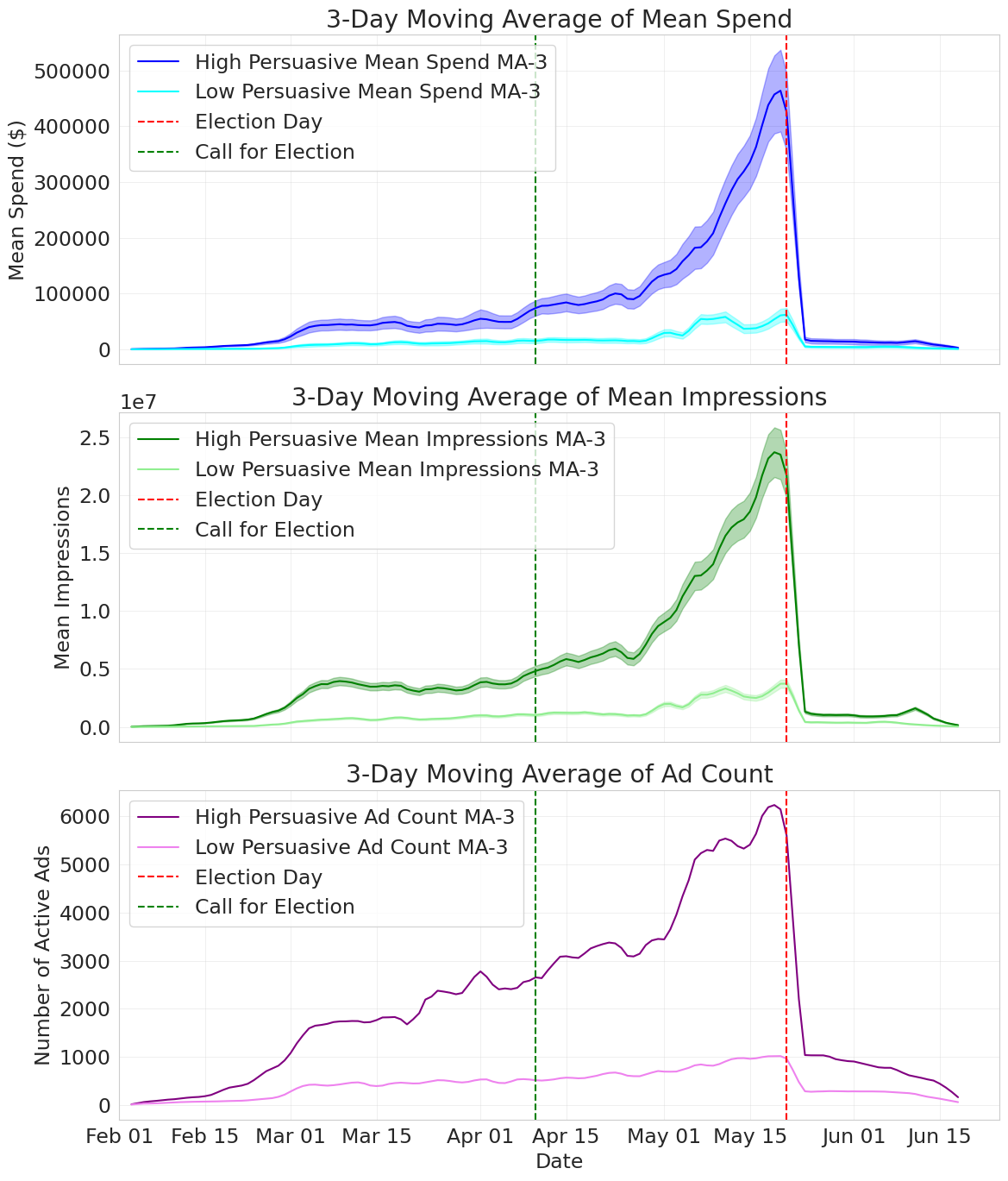}
  \caption{Time series of the total AUD spend, the number of impressions, and the number of unique ads for each day in the APA22 dataset. In the two upper panels the dashed area represents lower and upper bounds, with the solid line corresponding to the mean value. In all plots we use a 3-day moving average.} 
  \label{fig:ads_time_series}
\end{figure}

 Figure \ref{fig:ads_time_series} presents a time series analysis of the APA22 dataset, illustrating the daily trends of mean spend, mean impressions, and ad count for high and low persuasion ads. A noticeable divergence occurs after the call for election (April 10\textsuperscript{th}), where high persuasion ads exhibit a significant surge in all three metrics compared to their low persuasion counterparts. Specifically, the mean spend and impressions for high persuasion ads show a statistically significant increasing trend (Mann-Kendall test, \textit{P} < .001, $\alpha$ = .05) leading up to the election day (May 21\textsuperscript{st}), peaking sharply at 4.8 times the value measured on April 10\textsuperscript{th}, three days before the election. This is interesting given that Australia has a blackout period of three days before the election day which applies to broadcasters (TV and radio) but not to online services\footnote{\url{https://www.acma.gov.au/election-and-referendum-blackout-periods}}.
 
 In contrast, the low persuasion ads maintain a relatively steady trajectory with modest growth (2.2 times more). This pattern suggests a strategic amplification of high persuasion ad campaigns closer to the election date. There is a high correlation (Pearson \textit{R} = 0.99, \textit{P} <.001, $\alpha$ =.05) between aggregated daily spending and generated impressions. This is expected and indicates that the amount spent on ads generally determines the number of impressions generated. 
 This could also suggest minimal micro-targeting, as more targeted campaigns, which are typically more expensive, would likely result in fewer impressions for the same spending level. 
 The dataset acquired via the Meta Ad Library does not allow us to explore this further.


These insights, derived from our model predictions on the unlabeled APA22 data, provide an initial understanding of the use and impact of persuasive political advertising strategies. The findings underscore the potential of NLP models to extract subtle patterns from political ad content, paving the way for more extensive analyses in future research.

\subsection{Findings and Implications}
Study 2's examination of model adaptability and persuasive content analysis in the APA22 dataset yielded the following insights:

\paragraph{Domain Adaptability.} The model trained on the SemEval-23 dataset, primarily comprising news articles, showed limited effectiveness when applied to the social media political ads domain. This stresses the necessity for training models on domain-specific datasets to ensure their efficacy in real-world applications.
\paragraph{Persuasion Detection Performance.} Substantial improvements in accuracy and F1-score were observed when PPAsy-XLNet was retrained on the APA22 dataset, highlighting the benefits of domain-specific training. Nonetheless, the high efficiency of our model enables fast retraining reducing the need for domain adaptability.
\paragraph{Persuasive Content Dynamics.} High persuasive ads were associated with greater reach and financial investment, suggesting a strategic use of persuasive messaging to maximize impact.
\paragraph{Temporal Dynamics.} The escalation in advertisement spending and impressions for highly persuasive ads as the election approached indicates a strategic intensification of persuasive efforts to maximize influence before the voting day.
    


\section{Discussion and Conclusions}
In this work, we explored political advertising through the lens of persuasion techniques using our low-cost PPAsy-XLNet model and the APA22 dataset. Our model outperformed the SOTA in detecting persuasive text on the SemEval-23 Task 3 subtask 3 benchmark dataset, highlighting its efficiency and ability to handle multi-label classification.
We investigated domain adaptability and observed a performance dip when applying PPAsy-XLNet, initially trained on SemEval-23 data, to the social media political ads context. Retraining the model on a small number of manually labelled instances from the APA22 dataset improved accuracy and F1 scores.
Applying our model to the APA22 dataset unveiled insights into persuasion strategies in political advertising on social media, such as the subtle use of persuasive language, strategic ad spending, and targeted campaigns. This contributes to understanding how persuasion is implemented in the digital political sphere and aligns with the need for transparency and ethics in digital political campaigns.
Future work includes refining our classification strategy using a seq2seq prediction approach \citep{ostapuk2024follow}, expanding the APA22 dataset, and exploring a multimodal approach integrating image processing. These advancements could yield richer insights into the complex interplay of text and images in shaping public opinion and influencing electoral outcomes.

\section{Limitations}

\textbf{Study 1.} Our study encountered a challenge within the multi-label classification of the SemEval-23 Task 3 subtask3 due to the inaccessibility of the test set labels, which hindered our ability to perform a detailed error analysis. Furthermore, the discrepancy in the average number of persuasion techniques used in the development and test sets, as highlighted in our dataset overview (Table~\ref{tab:dataet_semeval_stats}), introduced additional challenges. The  persuasion techniques present in the development set did not closely mirror those in the test set, indicating that improvements or decreases in performance on the development set do not necessarily translate to similar changes in the test set. This limitation highlights the difficulty in calibrating our model's performance from development to test conditions, given the different intensity of the features in each set.

\textbf{Study 2.} The adaptation to a binary classification task in our analysis (due to the different annotation schemes used for the SemEval-23 and APA22 datasets) streamlined model training and evaluation but potentially restricted the granularity of our investigation into the depth of persuasive techniques in political ads. This meant that the diversity of persuasive strategies employed across various political campaigns might not have been fully observed. This limitation could impact the model ability to generalize across the wider spectrum of political advertising content encountered in different contexts.
Additionally, we experienced some limitations that inevitably arise when using data from the Meta Ad Library. The analysis was limited to political ads posted on Facebook, which may not fully represent the broader political campaigning strategies employed across different platforms. Our dataset comprised only ads explicitly flagged as political by the entity which posted it\footnote{It is legally required for political ads in Australia to have a "paid by" disclaimer}. This introduces potential biases, as some political ads might not have been flagged and were thus excluded from our data collection and analysis, while others could have been mis-flagged as political when they were not, which is a phenomenon that has been studied by \citet{10.1145/3442381.3450049}. Furthermore, Facebook reports only cover ranges of ad spend and reach, rather than precise figures, which can obfuscate the true scale and impact of individual ads. Lastly, the ad buying tool on Facebook allows advertisers to target audiences with a level of specificity that is not captured in the publicly available Ad Library data, concealing critical aspects of ad targeting strategies from our analysis.

\section{Ethics Statement}

The dataset collection for our study, referred to as \textit{APA22}, underwent ethics review by authors' institution IRB. Once the data was collected, the authors were  involved in the manual  annotation process.
The dataset we created will be made available with guidelines to prevent misuse and ensure research integrity. We recognize the potential implications of our work on the broader context of digital persuasion and are committed to promoting ethical standards in computational linguistic research about political communication.

\bibliography{custom}

\begin{thebibliography}{26}
\expandafter\ifx\csname natexlab\endcsname\relax\def\natexlab#1{#1}\fi

\bibitem[{Barrón-Cedeño et~al.(2019)Barrón-Cedeño, Jaradat, Da~San~Martino, and Nakov}]{barron2019}
Alberto Barrón-Cedeño, Israa Jaradat, Giovanni Da~San~Martino, and Preslav Nakov. 2019.
\newblock Proppy: A system to unmask propaganda in online news.
\newblock \emph{Information Processing \& Management}, 56(5):1849--1864.

\bibitem[{Boerboom(2020)}]{boerboom2020cambridge}
Carissa Boerboom. 2020.
\newblock Cambridge analytica: The scandal on data privacy.
\newblock \emph{Augustana Center for the Study of Ethics Essay Contest}.

\bibitem[{Chernyavskiy et~al.(2021)Chernyavskiy, Ilvovsky, and Nakov}]{chernyavskiy2021transformers}
Anton Chernyavskiy, Dmitry Ilvovsky, and Preslav Nakov. 2021.
\newblock Transformers:“the end of history” for natural language processing?
\newblock In \emph{Machine Learning and Knowledge Discovery in Databases. Research Track: European Conference, ECML PKDD 2021, Bilbao, Spain, September 13--17, 2021, Proceedings, Part III 21}, pages 677--693. Springer.

\bibitem[{Conneau et~al.(2019)Conneau, Khandelwal, Goyal, Chaudhary, Wenzek, Guzm{\'{a}}n, Grave, Ott, Zettlemoyer, and Stoyanov}]{DBLP:journals/corr/abs-1911-02116}
Alexis Conneau, Kartikay Khandelwal, Naman Goyal, Vishrav Chaudhary, Guillaume Wenzek, Francisco Guzm{\'{a}}n, Edouard Grave, Myle Ott, Luke Zettlemoyer, and Veselin Stoyanov. 2019.
\newblock \href {http://arxiv.org/abs/1911.02116} {Unsupervised cross-lingual representation learning at scale}.
\newblock \emph{CoRR}, abs/1911.02116.

\bibitem[{Cui et~al.(2019)Cui, Jia, Lin, Song, and Belongie}]{cui2019class}
Yin Cui, Menglin Jia, Tsung-Yi Lin, Yang Song, and Serge Belongie. 2019.
\newblock Class-balanced loss based on effective number of samples.
\newblock In \emph{Proceedings of the IEEE/CVF Conference on Computer Vision and Pattern Recognition}, pages 9268--9277.

\bibitem[{Da~San~Martino et~al.(2019)Da~San~Martino, Yu, Barrón-Cedeño, Petrov, and Nakov}]{dasmartino2019}
Giovanni Da~San~Martino, Seunghak Yu, Alberto Barrón-Cedeño, Rostislav Petrov, and Preslav Nakov. 2019.
\newblock Fine-grained analysis of propaganda in news articles.
\newblock In \emph{Proceedings of the 2019 Conference on Empirical Methods in Natural Language Processing and the 9th International Joint Conference on Natural Language Processing (EMNLP-IJCNLP)}, pages 5636--5646.

\bibitem[{Darmon et~al.(2021)Darmon, Bazzi, Howison, and Porter}]{darmon2021pull}
Alexandra~NM Darmon, Marya Bazzi, Sam~D Howison, and Mason~A Porter. 2021.
\newblock Pull out all the stops: Textual analysis via punctuation sequences.
\newblock \emph{European Journal of Applied Mathematics}, 32(6):1069--1105.

\bibitem[{Dimitrov et~al.(2021)Dimitrov, Ali, Shaar, Alam, Silvestri, Firooz, Nakov, and Martino}]{dimitrov2021detecting}
Dimitar Dimitrov, Bishr~Bin Ali, Shaden Shaar, Firoj Alam, Fabrizio Silvestri, Hamed Firooz, Preslav Nakov, and Giovanni Da~San Martino. 2021.
\newblock Detecting propaganda techniques in memes.
\newblock \emph{arXiv preprint arXiv:2109.08013}.

\bibitem[{Glowacki et~al.(2018)Glowacki, Narayanan, Maynard, Hirsch, Kollanyi, Neudert, Howard, Lederer, and Barash}]{glowacki2018news}
Monika Glowacki, Vidya Narayanan, Sam Maynard, Gustavo Hirsch, Bence Kollanyi, Lisa-Maria Neudert, Phil Howard, Thomas Lederer, and Vlad Barash. 2018.
\newblock News and political information consumption in mexico: Mapping the 2018 mexican presidential election on twitter and facebook.
\newblock \emph{The Computational Propaganda Project}.

\bibitem[{Habernal and Gurevych(2017)}]{habernal2017}
Ivan Habernal and Iryna Gurevych. 2017.
\newblock Argumentation mining in user-generated web discourse.
\newblock In \emph{Computational Linguistics}, volume~43, pages 125--179. MIT Press.

\bibitem[{Hristakieva et~al.(2022)Hristakieva, Cresci, Da~San~Martino, Conti, and Nakov}]{hristakieva2022spread}
Kristina Hristakieva, Stefano Cresci, Giovanni Da~San~Martino, Mauro Conti, and Preslav Nakov. 2022.
\newblock The spread of propaganda by coordinated communities on social media.
\newblock In \emph{Proceedings of the 14th ACM Web Science Conference 2022}, pages 191--201.

\bibitem[{Hromadka et~al.(2023)Hromadka, Smolen, Remis, Pecher, and Srba}]{Hromadka2023KinitveraAI}
Timo Hromadka, Timotej Smolen, Tomas Remis, Branislav Pecher, and Ivan Srba. 2023.
\newblock Kinitveraai at semeval-2023 task 3: Simple yet powerful multilingual fine-tuning for persuasion techniques detection.
\newblock \emph{arXiv preprint arXiv:2304.11924}.

\bibitem[{Lin et~al.(2017)Lin, Goyal, Girshick, He, and Doll{\'a}r}]{lin2017focal}
Tsung-Yi Lin, Priya Goyal, Ross Girshick, Kaiming He, and Piotr Doll{\'a}r. 2017.
\newblock Focal loss for dense object detection.
\newblock In \emph{Proceedings of the IEEE international conference on computer vision}, pages 2980--2988.

\bibitem[{Nakov et~al.(2021)Nakov, Alam, Shaar, Da~San~Martino, and Zhang}]{nakov2021covid}
Preslav Nakov, Firoj Alam, Shaden Shaar, Giovanni Da~San~Martino, and Yifan Zhang. 2021.
\newblock Covid-19 in bulgarian social media: Factuality, harmfulness, propaganda, and framing.
\newblock In \emph{Proceedings of the International Conference on Recent Advances in Natural Language Processing (RANLP 2021)}, pages 997--1009.

\bibitem[{Ostapuk et~al.(2024)Ostapuk, Audiffren, Dolamic, Mermoud, and Cudr{\'e}-Mauroux}]{ostapuk2024follow}
Natalia Ostapuk, Julien Audiffren, Ljiljana Dolamic, Alain Mermoud, and Philippe Cudr{\'e}-Mauroux. 2024.
\newblock Follow the path: Hierarchy-aware extreme multi-label completion for semantic text tagging.
\newblock In \emph{Proceedings of the ACM on Web Conference 2024}, pages 2094--2105.

\bibitem[{Perloff(2017)}]{Perloff2017Dynamics}
R.M. Perloff. 2017.
\newblock \href {https://doi.org/10.4324/9781315657714} {\emph{The Dynamics of Persuasion: Communication and Attitudes in the Twenty-First Century}}, 6th edition.
\newblock Routledge.

\bibitem[{Piskorski et~al.(2023{\natexlab{a}})Piskorski, Stefanovitch, Da~San~Martino, and Nakov}]{piskorski2023semeval}
Jakub Piskorski, Nicolas Stefanovitch, Giovanni Da~San~Martino, and Preslav Nakov. 2023{\natexlab{a}}.
\newblock Semeval-2023 task 3: Detecting the category, the framing, and the persuasion techniques in online news in a multi-lingual setup.
\newblock In \emph{Proceedings of the 17th International Workshop on Semantic Evaluation (SemEval-2023)}, pages 2343--2361.

\bibitem[{Piskorski et~al.(2023{\natexlab{b}})Piskorski, Stefanovitch, Nikolaidis, Da~San~Martino, and Nakov}]{piskorski2023multilingual}
Jakub Piskorski, Nicolas Stefanovitch, Nikolaos Nikolaidis, Giovanni Da~San~Martino, and Preslav Nakov. 2023{\natexlab{b}}.
\newblock Multilingual multifaceted understanding of online news in terms of genre, framing, and persuasion techniques.
\newblock In \emph{Proceedings of the 61st Annual Meeting of the Association for Computational Linguistics (Volume 1: Long Papers)}, pages 3001--3022.

\bibitem[{Purificato and Navigli(2023)}]{purificato2023}
Antonio Purificato and Roberto Navigli. 2023.
\newblock Apatt at semeval-2023 task 3: The sapienza nlp system for ensemble-based multilingual propaganda detection.
\newblock In \emph{Proceedings of the The 17th International Workshop on Semantic Evaluation (SemEval-2023)}, pages 382--388.

\bibitem[{Rashkin et~al.(2017)Rashkin, Choi, Jang, Volkova, and Choi}]{rashkin2017}
Hannah Rashkin, Eunsol Choi, Jin~Yea Jang, Svitlana Volkova, and Yejin Choi. 2017.
\newblock Truth of varying shades: Analyzing language in fake news and political fact-checking.
\newblock In \emph{Proceedings of the 2017 Conference on Empirical Methods in Natural Language Processing}, pages 2931--2937.

\bibitem[{Salman et~al.(2023)Salman, Hanif, Shehata, and Nakov}]{salman2023detecting}
Muhammad~Umar Salman, Asif Hanif, Shady Shehata, and Preslav Nakov. 2023.
\newblock Detecting propaganda techniques in code-switched social media text.
\newblock \emph{arXiv preprint arXiv:2305.14534}.

\bibitem[{Sosnovik and Goga(2021)}]{10.1145/3442381.3450049}
Vera Sosnovik and Oana Goga. 2021.
\newblock \href {https://doi.org/10.1145/3442381.3450049} {Understanding the complexity of detecting political ads}.
\newblock In \emph{Proceedings of the Web Conference 2021}, WWW '21, page 2002–2013, New York, NY, USA. Association for Computing Machinery.

\bibitem[{Tardaguila et~al.(2018)Tardaguila, Benevenuto, and Ortellado}]{Tardaguila2018FakeNews}
Cristina Tardaguila, Fabrício Benevenuto, and Pablo Ortellado. 2018.
\newblock Fake news is poisoning brazilian politics. whatsapp can stop it.
\newblock \url{https://www.nytimes.com/2018/10/17/opinion/brazil-election-fake-news-whatsapp.html}.
\newblock Accessed: 15 Feb 2024.

\bibitem[{Teimas and Saias(2023)}]{teimas2023detecting}
R{\'u}ben Teimas and Jos{\'e} Saias. 2023.
\newblock Detecting persuasion attempts on social networks: Unearthing the potential of loss functions and text pre-processing in imbalanced data settings.
\newblock \emph{Electronics}, 12(21):4447.

\bibitem[{Yang et~al.(2019)Yang, Dai, Yang, Carbonell, Salakhutdinov, and Le}]{DBLP:journals/corr/abs-1906-08237}
Zhilin Yang, Zihang Dai, Yiming Yang, Jaime~G. Carbonell, Ruslan Salakhutdinov, and Quoc~V. Le. 2019.
\newblock \href {http://arxiv.org/abs/1906.08237} {Xlnet: Generalized autoregressive pretraining for language understanding}.
\newblock \emph{CoRR}, abs/1906.08237.

\bibitem[{Yu et~al.(2021)Yu, Martino, Mohtarami, Glass, and Nakov}]{yu2021interpretable}
Seunghak Yu, Giovanni Da~San Martino, Mitra Mohtarami, James Glass, and Preslav Nakov. 2021.
\newblock Interpretable propaganda detection in news articles.
\newblock \emph{arXiv preprint arXiv:2108.12802}.

\end{thebibliography}
\bibliographystyle{acl_natbib}

\newpage
\appendix

\label{sec:semeval_annotation}

\begin{table*}[ht]
\centering
\resizebox{\textwidth}{!}{%
\begin{tabular}{l|rrr|rrr|rrr|rr}
\hline
\textbf{Language} & \multicolumn{3}{c|}{\textbf{Train}} & \multicolumn{3}{c|}{\textbf{Dev}} & \multicolumn{3}{c|}{\textbf{Test}} & \multicolumn{2}{c}{\textbf{Total}} \\
 & \textbf{\#DOC} & \textbf{\#CHAR} & \textbf{AVG pt} & \textbf{\#DOC} & \textbf{\#CHAR} & \textbf{AVG pt} & \textbf{\#DOC} & \textbf{\#CHAR} & \textbf{AVG pt} & \textbf{\#DOC} & \textbf{\#WORD} \\
\hline
EN & 446 & 2.43M & 16.1 & 90 & 403K & 20.0 & 54 & 228K & 32.9 & 590 & 469K \\
FR & 158 & 737K & 35.4 & 53 & 222K & 29.9 & 50 & 181K & 33.6 & 261 & 153K \\
DE & 132 & 581K & 34.1 & 45 & 171K & 27.5 & 50 & 259K & 38.1 & 227 & 104K \\
IT & 227 & 927K & 26.6 & 76 & 287K & 25.4 & 61 & 245K & 38.5 & 364 & 186K \\
PL & 145 & 765K & 19.6 & 49 & 264K & 20.1 & 47 & 349K & 31.7 & 241 & 144K \\
RU & 143 & 590K & 23.8 & 48 & 163K & 15.4 & 72 & 161K & 13.1 & 263 & 104K \\
GE & - & - & - & - & - & - & 29 & 46K & 7.5 & 29 & - \\
GR & - & - & - & - & - & - & 64 & 248K & 10.8 & 64 & - \\
SP & - & - & - & - & - & - & 30 & 109K & 18.2 & 30 & - \\
\hline
\end{tabular}
}
\caption{Overview of SemEval 2023 dataset across various languages, showcasing document counts, character totals, with average persuasion techniques identified per document (AVG pt) for training (Train), development (Dev), and testing (Test) sets. Georgian (GE), Greek (GR), and Spanish (SP), introduced as surprise languages, are represented solely within the Test.  The `Total' column summarizes the total number of documents (\#DOC) and words (\#WORD) for each language across all sets.}
\label{tab:dataet_semeval_stats}
\end{table*}

\section{Text Pre-processing}
\label{sec:tf-idf}

In Study 2, we applied a series of pre-processing steps to the text data obtained from the ads to facilitate linguistic analysis using TF-IDF and n-grams. The pre-processing pipeline included converting the text to lowercase, removing punctuation, emojis, and links, and stripping whitespaces. Additionally, we filtered out stopwords to focus on more semantically meaningful terms, and tokenized the text into individual units. These pre-processing steps were useful for reducing noise and improving our linguistic analysis.

\section{TF-IDF}

\begin{table*}[ht]
\centering
\begin{tabular}{lcc}
\hline
\textbf{} & \textbf{High Persuasion} & \textbf{Low Persuasion} \\
\hline
\multirow{10}{*}{TF-IDF} 
& labor & community \\
& vote & local \\
& government & support \\
& morrison & team \\
& future & labor \\
& need & time \\
& better & nsw \\
& australia & join \\
& community & service \\
& plan & vote \\
\hline
\end{tabular}
\caption{Ten highest TF-IDF words for high and low persuasive ads respectively}
\label{tab:tf-idf}
\end{table*}

For the analysis of the textual component of the ads, we employed the Term Frequency-Inverse Document Frequency (TF-IDF) technique to assess the importance of words within the corpus. This method measures the frequency of a word in a single advertisement and also accounts for the popularity of the word across all ads.

To ascertain a global significance score for each word across all ads, we computed the average TF-IDF score for each term across the dataset. This average score represents the overall importance of each word within the entire corpus.
The results of our TF-IDF analysis are summarized by listing the  10 words with the highest average TF-IDF scores for both high and low persuasion ads (see Table \ref{tab:tf-idf}).

\section{Bi-gram}

Our methodology also involves the extraction and evaluation of n-grams from text data to uncover frequently occurring  word patterns. These n-grams are sorted to ensure that identical phrases with words in different orders are counted as the same n-gram. For instance, ('word1', 'word2') and ('word2', 'word1') are counted as a single bi-gram. We chose bi-grams because they provide an optimal balance between capturing meaningful lexical relationships and ensuring a sufficient frequency of occurrences. Bi-grams allow us to observe the most  significant word pairs, providing insights into the structure of persuasive language commonly used in political advertising.

While word co-occurrence matrices provide useful insights, particularly in identifying broader semantic connections and themes within large text corpora, they can sometimes reduce the focus on direct linguistic interactions, which are crucial in advertising. In contrast, N-gram analysis offers a more targeted approach by directly highlighting the sequences of words that frequently occur together. This provides a clearer view of common linguistic patterns and potentially persuasive language strategies used in online advertisements.

\begin{table*}[ht]
\centering
\begin{tabular}{lcc}
\hline
\textbf{} & \textbf{High Persuasion} & \textbf{Low Persuasion} \\
\hline
\multirow{10}{*}{Bi-grams} 
& morrison scott & election federal \\
& government labor & job vacancy \\
& change climate & melbourne vic \\
& better future & government labor \\
& aged care & strait torres \\
& cost living & islander strait \\
& petition sign & aboriginal torres \\
& economy strong & business small \\
& future stronger & health mental \\
& government morrison & candidate labor \\
\hline
\end{tabular}
\caption{Top ten bi-grams for high and low persuasive ads respectively}
\label{tab:bi-grams}
\end{table*}

\begin{figure*}[ht]
    \centering
    \begin{subfigure}{.45\textwidth}  
        \centering
        \includegraphics[width=\linewidth]{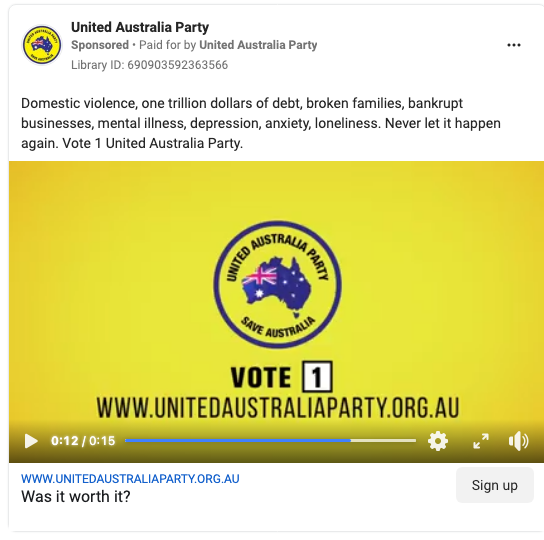} 
        \caption{High persuasion ad by the United Australia Party utilizing fear-based messaging and strong emotive language to provoke an immediate emotional response. It claims severe societal consequences if the party is not voted for, potentially enhancing its persuasive impact.}
        \label{fig:high-persuasive-ad}
    \end{subfigure}%
    \hfill 
    \begin{subfigure}{.45\textwidth}  
        \centering
        \includegraphics[width=\linewidth]{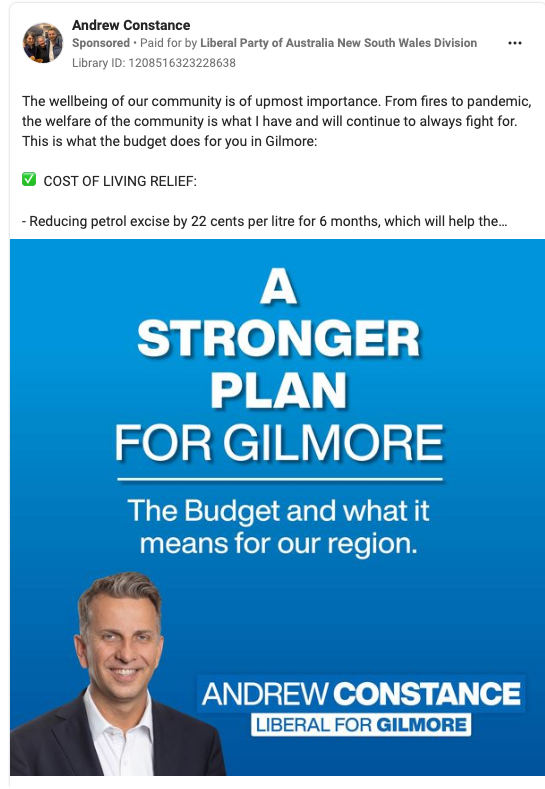} 
        \caption{Ad by the Liberal Party of Australia, employing a more informational and optimistic approach. It focuses on practical benefits and community well-being, which is less emotionally charged compared to the United Australia Party ad.}
        \label{fig:low-persuasive-ad}
    \end{subfigure}
    \caption{Comparison of two political advertisements with differing persuasive intensities.}
\end{figure*}

\section{Computational Resources}
\label{sec:comp_resources}
The experiments were performed using a Google Colab environment with an NVIDIA A100 GPU (40 GB of GPU RAM) and 83.5 GB of system RAM.
\end{document}